\documentclass[10pt,twocolumn,letterpaper]{article}

\usepackage[pagenumbers]{main}

\usepackage[table]{xcolor}
\usepackage{microtype}
\usepackage{graphicx}
\usepackage{booktabs}
\usepackage{amsmath}
\usepackage{amssymb}
\usepackage{mathtools}
\usepackage{amsthm}

\usepackage{color}
\usepackage{enumitem}
\usepackage{multirow}
\usepackage{makecell}

\usepackage{algorithmic}
\usepackage{algorithm}
\usepackage{etoolbox,siunitx}

\usepackage{adjustbox}
\usepackage{abstract}
\usepackage{caption}

\definecolor{lblue}{rgb}{0.21,0.49,0.74}

\usepackage[pagebackref,breaklinks,colorlinks,citecolor=lblue]{hyperref}

\title{DINO-SD: Champion Solution for ICRA 2024 RoboDepth Challenge} 

\author{
  Yifan~Mao\thanks{Equal contribution.}\hspace{12pt}
  Ming~Li\footnotemark[1]\hspace{12pt}
  Jian~Liu\footnotemark[1]\hspace{12pt}
  Jiayang~Liu\hspace{12pt}
  Zihan~Qin\hspace{12pt}
  Chunxi~Chu \\
  Jialei~Xu\hspace{12pt}
  WenBo~Zhao\hspace{12pt}
  Junjun~Jiang\hspace{12pt}
  Xianming~Liu\thanks{Corresponding author.}\\
  \large{Harbin Institute of Technology} \\
}

\begin{document}

\maketitle
\renewcommand{\thefootnote}{\fnsymbol{footnote}}


\begin{abstract}

\textit{
Surround-view depth estimation is a crucial task aims to acquire the depth maps of the surrounding views. It has many applications in real world scenarios such as autonomous driving, AR/VR and 3D reconstruction, etc. However, given that most of the data in the autonomous driving dataset is collected in daytime scenarios, this leads to poor depth model performance in the face of out-of-distribution(OoD) data. While some works try to improve the robustness of depth model under OoD data, these methods either require additional training data or lake generalizability. In this report, we introduce the DINO-SD, a novel surround-view depth estimation model. Our DINO-SD does not need additional data and has strong robustness. Our DINO-SD get the best performance in the track4 of ICRA 2024 RoboDepth Challenge.
}
\end{abstract}

\section{Overview}

Depth estimation, which aims to estimate the distance of every point in the image, is a crucial task in 3D vision with important applications such as autonomous driving\cite{mgnet}, augmented reality\cite{aug_real}, virtual reality\cite{panodepth}, and 3D reconstruction\cite{mobile3drecon}. Compared to acquiring depth using depth sensors such as LiDAR, estimating depth from images can effectively reduce hardware costs and produce dense depth map. This makes depth estimation algorithms the primary choice in these fields.

Depth estimation tasks can be categorized into monocular depth estimation and multi-view depth estimation according to the number of cameras used. Monocular depth estimation is inherently limited by its reliance on a single viewpoint, which compromises its robustness. In contrast, multi-view depth estimation provides a comprehensive $360^{\circ}$ view of the surroundings, which is able to more accurate estimation of depth and robust to changes in scene geometry. As sensor technologies advance and manufacturing costs decrease,  multi-view depth estimation has progressively replaced monocular depth estimation as the industry standard. Notable works in this field include MVSNet\cite{mvsnet}, SurroundDepth\cite{surrounddepth} and S3Depth\cite{S3Depth}.

However, existing multi-view depth estimation methods still do not perform satisfactorily in real-world scenarios. The main reason is that real-world sensor data often contains corruptions, such as adverse weather and sensor noise, and most autonomous driving training datasets primarily consist of clean data.  
Existing methods \cite{surrounddepth, S3Depth} lack robustness to noise. Several studies have focused on enhancing the robustness of depth estimation\cite{r4dyn, MS2, md4all, SSD} through the use of additional training data in different scenes. However, the introduction of additional data still cannot cover all situations in real-world scenarios. Given the prohibitive cost of acquiring large volumes of corrupted data, and considering that expanding training datasets may not encompass all real-world corruptions, developing a robust multi-view depth estimation model capable of performing well on out-of-distribution (OoD) data is imperative.

Inspiration from the Depth Anything\cite{depthanything} framework, we introduce DINO-SD, a novel approach aimed at improving the robustness of surround-view depth estimation models. This framework is designed to handle a variety of environmental conditions and sensor imperfections, enhancing the reliability of depth estimation in autonomous driving and other critical applications.




\section{Related Work}
\subsection{Multi-view Depth Estimation}
Multi-view depth estimation aims at estimating the depth of the scene using multi-view information. It can be categorized into stereo matching, multi-view stereo and surround-view depth estimation. 

Stereo matching utilizes the left and right camera images to predict the depth. The stereo matching algorithm recovers the depth information of the scene through the basic principles of multi-view geometry.
Multi-view stereo matching improves depth estimation model performance by computing the differentiable cost volume of multiple viewpoints and implementing feature fusion to aggregate feature information from multiple viewpoints.
MVSNet\cite{mvsnet} is the first work to use multiple viewpoints cost volume to estimate the scene depth. 
Surround-view depth estimation is also a kind of multi-view depth estimation which takes $360^{\circ}$ surrounding views as input. SurroundDepth\cite{surrounddepth} first uses self-attention to process surrounding view images and get the depth maps of surrounding views. S3Depth\cite{S3Depth} changes the self attention to adjacent-view cross attention and achieves better performance.

\subsection{Robust Depth Estimation}
Although depth estimation model performance is greatly improved in daytime scenes, it suffers severely when confronted with OoD data. Some works have wxplored robust depth estimation, and these can be divided into two categories: data-driven robust depth estimation and model-driven robust depth estimation. 

Model-driven robust depth estimation achieves robust depth estimation by introducing new modules or modifying depth model architecture.  
DeFeat-Net\cite{DeFeat-Net} proposes a unified framework aims at improving depth model performance under darkness.
RNW\cite{RNW} uses image enhancement technique and adversarial training to enhance model performance under darkness. 
WSGD\cite{WSGD} estimates the flow maps and models light changes between adjacent frames to enhance model performance under darkness.
The problem with model-driven robust depth estimation is that it is poorly migratable and can only be adapted to a single adverse scene, not to multiple OoD data.

Data-driven robust depth estimation achieves robust depth estimation by introducing other modalities into model training or using the domain adaption to scale up the training dataset. 
DEISR\cite{DEISR} uses sparse Radar data to enable depth estimation under adverse conditions.
R4Dyn\cite{r4dyn} uses sparse radar data as weak supervision during training. 
DET\cite{DET} uses thermal images to estimate depth.
ADDS\cite{ADDS} uses different encoders to extract invariant and private features of different domains.
ITDFA\cite{ITDFA} utilizes CycleGAN\cite{CycleGAN} to translate images into other domains and uses domain adaption method to improve the model robustness.
Md4all\cite{md4all} achieves robust depth estimation by not distinguishing images in standard and challenging conditions, and training the model using knowledge distillation.
The problem with methods using other modalities is that the high cost of data collection equipment and manual data labeling limits the development of these methods. The problem with domain adaption is that it needs a translation model to convert normal distribution data into OoD data which requires training translation model and generate training data. 

Current robust depth estimation methods lack a generalized model architecture that can be adapted to a wide range of out-of-distribution data, and also rely on additional modal data or artificial data training. How to achieve robust depth estimation without using out-of-distribution data or additional modal training is a question worth exploring.

\section{Technical Approach}
\subsection{Overview}

Given 6 surrounding views $I_s \in \mathbb{R}^{6 \times 3 \times H \times W}$.
The goal of the proposed DINO-SD is to output 6 corresponding depth maps $D_s \in \mathbb{R}^{6 \times 1 \times H \times W}$. As shown in Fig.~\ref{fig:DINO-SD}, the proposed DINO-SD encompasses three principal phases: feature extracting, fuse and decoding, depth estimation. In the following, we will introduce the three phases in details.



\subsection{DINO-SD}
\begin{figure*}[h]
    \centering
    \includegraphics[width=0.76\linewidth, height=0.36\linewidth]{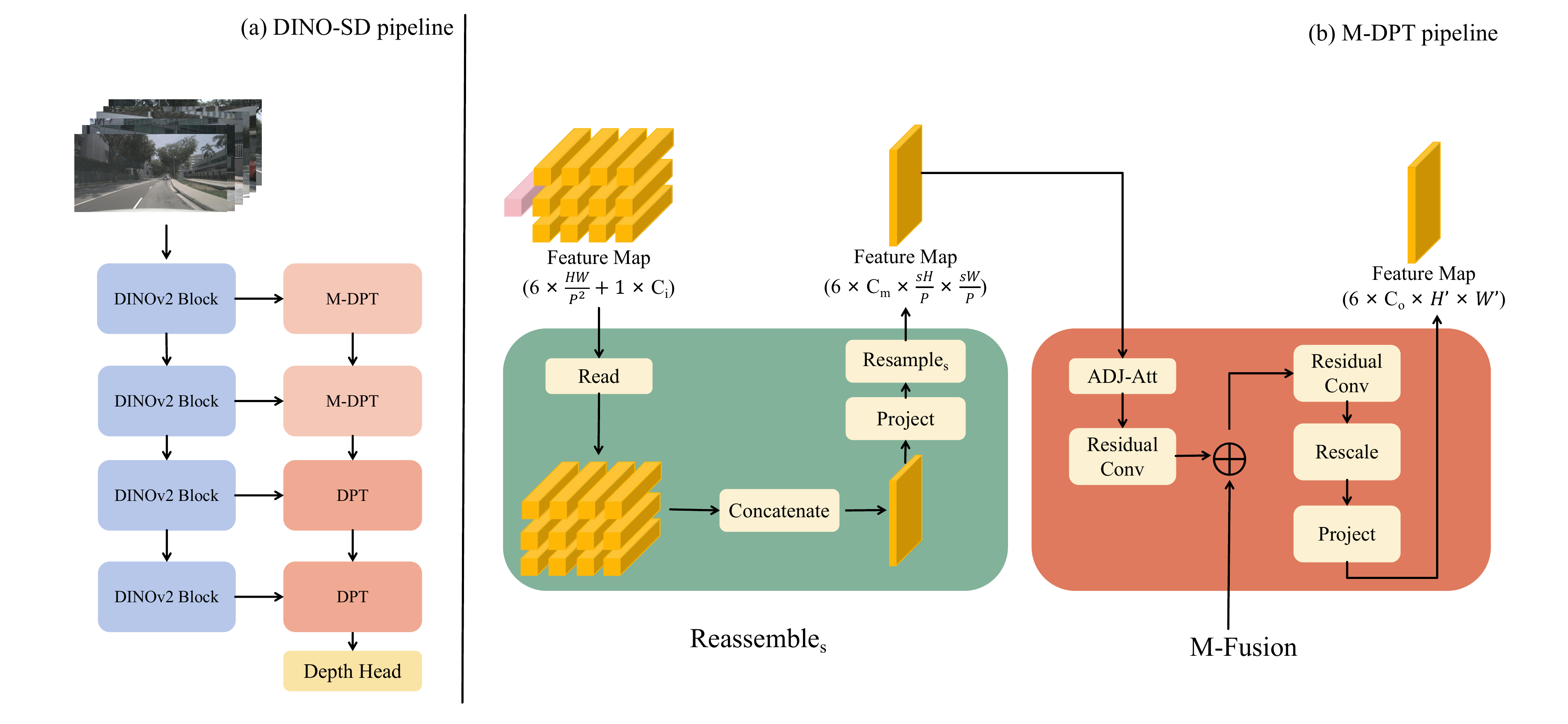}
    \caption{\textbf{Our DINO-SD model}: Our DINO-SD model use the pretrained DINOv2 as encoder, M-DPT and DPT as decoder. }
    \label{fig:DINO-SD}
\end{figure*}

Our DINO-SD uses the pretrained DINOv2\cite{dinov2} as encoder, and M-DPT and DPT \cite{DPT} as decoders. The reason we use Dinov2 is that Dinov2 can extract robust image features compared with other encoders. It helps to improve the model performance when processing the OoD data. 

Furthermore, we choose DPT\cite{DPT} as our decoder. We also modify the structure of DPT to adapt for surround-view depth estimation and propose the Multiview-DPT (M-DPT) as shown in figure\ref{fig:DINO-SD}. We introduce the adjacent-view attention into DPT. 

In previous surround-view depth estimation task, SurroundDepth\cite{surrounddepth} uses cross-view self attention while S3Depth\cite{S3Depth} uses adjacent-view cross attention. 

Let $F_i \in \mathbb{R}^{N \times C \times \frac{H}{n} \times \frac{W}{n}}, i = 1,2,3,4,5,6$ be the feature maps obtained from \textit{i}-th view, where $H$ and $W$ indicate the height and width of the input images, $N$ represents the batchsize, and $C$ stands for the dimensions of the feature map. For self attention, the feature maps $F_i$ are concated and reshaped into $F \in \mathbb{R}^{N \times \frac{6HW}{n^2} \times C}$ and then used to compute the \textit{Q}, \textit{K}, \textit{V} from $F$. The formular \ref{equ:self-att} shows the calculation process.
\begin{equation}
    \begin{aligned}
       Q &= W_QF\\
       K &= W_KF\\
       V &= W_VF\\
       F &= sofrmax(\frac{Q^{T}K}{\sqrt{C}})V 
    \end{aligned}    
    \label{equ:self-att}
\end{equation}
For adjacent-view cross attention, \textit{K} and \textit{V} are computed from the adjacent-view feature maps $F_j, j \in (i-1, i+1)$. The feature maps $F_i$ are shaped into $F_i \in \mathbb{R}^{N \times \frac{HW}{n^2} \times C}$ and then used to computed the \textit{Q}, \textit{K}, and \textit{V}. The formular \ref{equ:cross-att} shows the calculation process.
\begin{equation}
    \begin{aligned}
        Q &= W_QF_i\\
        K &= W_KF_j\\
        V &= W_VF_j\\
        F_i &= softmax(\frac{Q^{T}K}{\sqrt{C}})V
    \end{aligned}
    \label{equ:cross-att}
\end{equation}

S3Depth's performance is better than SurroundDepth, so we use the adjcent-view cross attention. We also try the cross-view self attention but the results show that adjacent-view cross attention is better than cross-view self attention. More details can be found in section \ref{sec:exp}.
To introduce self attention or cross attention into DPT, we perform the self attention or cross attention operation before the feature maps are fed into the Fusion module of DPT. More details for DPT structure can be found in \cite{DPT}.

Our depth head is very simple, containing only two convolutional layers and a sigmoid head.

\subsection{Training Pipeline}
\begin{figure*}[h]
    \centering
    \begin{subfigure}{0.52\linewidth}
       \includegraphics[width=\linewidth]{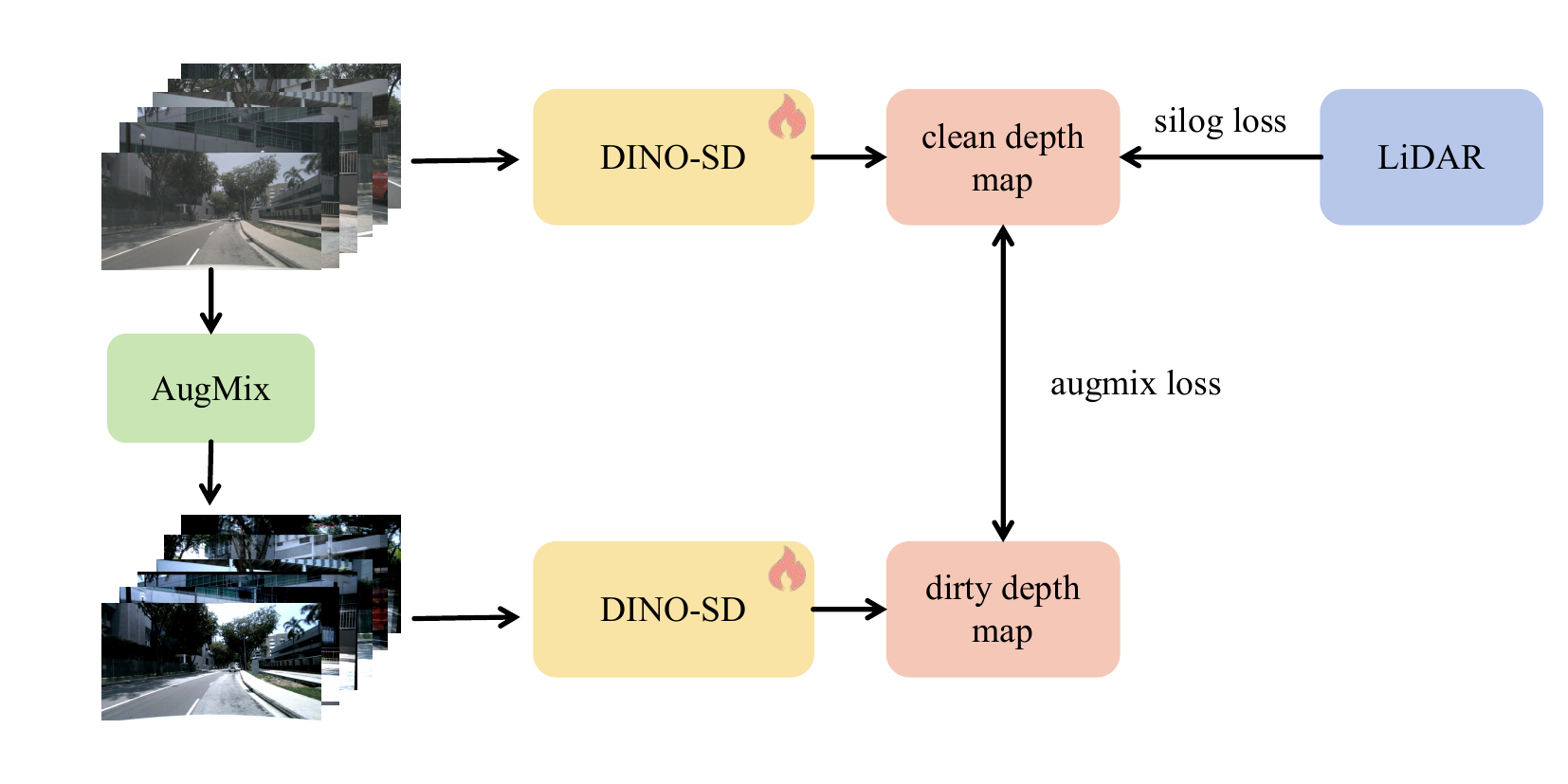}
        \caption{Training pipeline.}
        \label{fig:training-pipeline} 
    \end{subfigure}
    \hfill
    \begin{subfigure}{0.42\linewidth}
        \includegraphics[width=\linewidth]{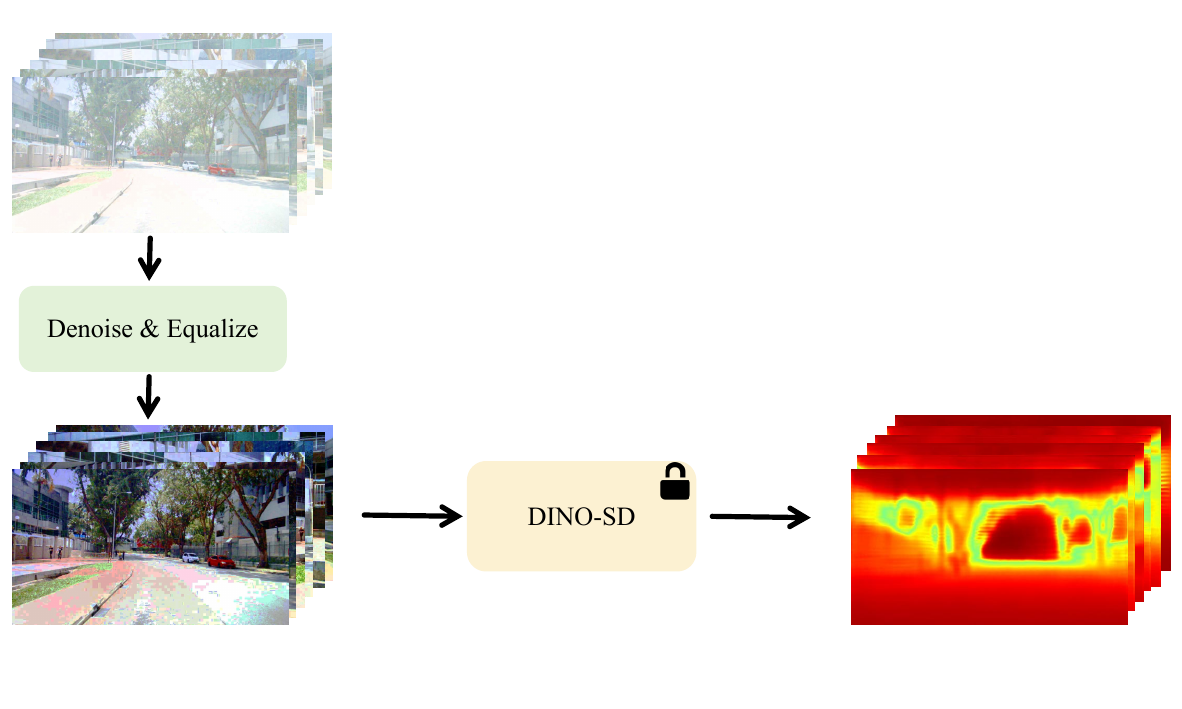}
        \caption{Testing pipeline.}
        \label{fig:testing-pipeline}
    \end{subfigure}
    \caption{Our training and testing pipeline.}
\end{figure*}

Our model training pipeline is shown in the figure \ref{fig:training-pipeline}. For surrounding-view images $I_s \in \mathbb{R}^{6 \times 3 \times H \times W}$, we first use DINOv2 to extract the feature maps $F \in \mathbb{R}^{6 \times \frac{HW}{n^2} \times C}$ and then we use M-DPT and DPT to decode the feature maps. Finally, we use a depth head to get the depth map $D \in \mathbb{R}^{6 \times 1 \times H \times W}$. We use the LiDAR ground truth to provide the supervision for depth map. We adopt the silog loss to depth supervision:
\begin{equation}
    \begin{aligned}
        L_{silog} = \frac{1}{n} \sum_{i} d_i^2 - \frac{\lambda}{n^2} (\sum_{i} d_i)^2,
    \end{aligned}
\end{equation}
where $d_i = \log y_i - \log y_i^*$, $y_i$ represents the predicted depth of $i$-th pixel and $y_i^*$ represents the ground truth depth of $i$-th pixel. We set $\lambda = 0.85$. 

Furthermore, we use the AugMix loss to make the model be capable to process OoD data. AugMix data augmentation method\cite{augmix} which mixes the augmented images. We first use AugMix to process the surrounding-view images $I_s$ and get the augmented images $I_a$. Then, we use our DINO-SD to estimate the depth maps $D_a$ of the augmented images $I_a$, and use the AugMix loss to make $D_s$ and $D_a$ have similar distributions. In our experiments, we perform two separate AugMix operations on $I_s$ to obtain $D_{a1}$ and $D_{a2}$, and calculate the difference between the $D_s$, $D_{a1}$, $D_{a2}$ distributions by JS divergence:
\begin{equation}
    \begin{aligned}
        D_{mix} &= \frac{1}{3}(D_s + D_{a1} + D_{a2}), \\
        L_{AugMix} &= \frac{1}{3}(KL(D_s||D_{mix}) + KL(D_{a1}||D_{mix})\\ & \quad+ KL(D_{a2}||D_{mix})),
    \end{aligned}
\end{equation}
where $D_{mix}$ represents the mixed depth map and $KL$ represents the KL divergence.

We also use the smooth loss to maintain the depth map consistency:
\begin{equation}
\begin{aligned}
L_{smooth} = \sum_i |\partial_x d_i^*|e^{-|\partial_x I_i|} + |\partial_y d_i^*|e^{-|\partial_y I_i|},
\end{aligned}
\end{equation}
where $d_i^* = d_i / \overline{d_i}$.

Our loss is the combination of silog loss, augmix loss and smooth loss:
\begin{equation}
\begin{aligned}
    L = L_{silog} + \alpha L_{smooth} + \beta L_{AugMix},
\end{aligned}
\end{equation}
we set $\alpha = 10^{-3}$ and $\beta = 10^{-2}$.

\subsection{Testing Pipeline}
Our testing pipeline is shown in figure\ref{fig:testing-pipeline}. For surrounding-view OoD images, we perform image denoise and equalize operations on the OoD images and then input into our trained DINO-SD. We adopt donoho \textit{et al}. image denoise method\cite{denoise}. We did not use model ensemble methods to improve the performance of our method.

\section{Experiments}
\label{sec:exp}

\begin{table*}[h]
\centering
\captionsetup{
  justification=justified,
  width=1\textwidth,
   margin={0pt,0pt}
}
\caption{Ablation results of DinoSurDepth on the RoboDrive cimpetition leaderboard.The symbol $\times$ indicates that the module was not used, the symbol $\checkmark$ indicates that the module was used, and the best results are highlighted in \textbf{bold}.}
\label{tab:ablation_result}
\begin{adjustbox}{width=1\textwidth}
\begin{tabular}{@{}lccccc ccc c c@{}}
\toprule
Method        &attention  &denoise &equalize            & Abs Rel$\downarrow$ & Sq Rel$\downarrow$ & RMSE$\downarrow$ & log RMSE$\downarrow$ & $a1\uparrow$ & $a2\uparrow$ & $a3\uparrow$ \\ \midrule
SurroundDepth\newline(Baseline)  &self attention    & $\times$ & $\times$ & 0.3039 & 3.0596 & 8.5285 & 0.4003 & 0.5439 & 0.7839 & 0.8911 \\
DINO-SD       &$\times$  &$\times$ &$\times$      & 0.2654 & 2.5803 & 8.2702 & 0.3738 & 0.5873 & 0.8220 & 0.9128 \\
DINO-SD       &self attention &$\times$  &$\times$       & 0.2162 & 1.9451 & 7.6721 & 0.3289 & 0.6702 & 0.8512 & 0.9279 \\
DINO-SD       &self attention &$\checkmark$ &$\times$  & 0.2074 & 1.8087 & 7.4020 & 0.3145 & 0.6887 & 0.8588 & 0.9340 \\
DINO-SD       &self attention &$\times$  &$\checkmark$      & 0.2078 & 1.7107 & 7.0786 & 0.3091 & 0.6844 & 0.8611 & 0.9370 \\
DINO-SD       &self attention &$\checkmark$  &$\checkmark$ & 0.2052 & 1.7246 & 7.1974 & 0.3075 & 0.6900 & 0.8651 & 0.9395 \\
DINO-SD       &adjacent-view cross attention &$\checkmark$  &$\checkmark$ & \textbf{0.1870} & \textbf{1.4683} & \textbf{6.2365} & \textbf{0.2760} & \textbf{0.7339} & \textbf{0.8952} & \textbf{0.9531} \\
\bottomrule
\end{tabular}
\end{adjustbox}
\end{table*}

\noindent\textbf{Implementation Details} The DINO-SD framework is implemented using PyTorch. It utilizes four NVIDIA GTX 3090 GPUs for model training, each configured as a batch size of 1x6 (for six views). On the leaderboard, our method secured first place across all six metrics. We use the DINOv2 \cite{dinov2} as the backbone, which is pretrained on a massive dataset named LVD-142M, comprising 142 million images. This dataset is assembled from ImageNet-22k, the training split of ImageNet-1k, Google Landmarks, and several fine-grained datasets. In the DINO-SD model, we use the last four blocks of the DINOv2 encoder to extract image features, the M-DPT and DPT to decode feature maps. For the third-to-last and fourth-to-last blocks, we processed the features using M-DPT, with the resample scale set to 1 and 0.5 respectively. For the penultimate and penultimate blocks, we processed the features using DPT with the resample scale set to 2 and 4, respectively. The learning rate for the encoder is 5e-6 and the learning rate for the decoder is 2e-5. We employed the CosineAnnealingWarmRestarts method for our model's optimizer. This scheduler adjusts the learning rate following a cosine annealing pattern, periodically resetting to enhance convergence. We spent 18 hours training for 5 epochs, but ultimately found that the weights from the first epoch achieved the best results on the test set (corrupted images). We believe this was mainly due to the following reasons: Firstly, our batch size was relatively large (4 GPUs × 6 views), which led to the model converging quickly; Secondly, there was a significant distribution shift between the corrupted images and clean images, and learning too much from the clean images easily led to overfitting. 

\noindent\textbf{Comparative Study} The benchmark uses the original depth of the corruption images as ground truth for evaluation purposes. Corruptions are simulated through algorithms encompassing 18 types of corruptions, namely: darkness, brightness, defocus blur, contrast, JPEG compression, impulse noise, motion blur, snow, zoom blur, frost, pixelation, color, quantization, elastic transformation, Gaussian noise, fog, ISO noise, shot noise, and glass blur. Table \ref{tab:performance_comparison} compares the model's performance with that of other teams on the RoboDrive competition leaderboard.

\begin{table}[H]
\centering
\captionsetup{
  justification=justified,
  singlelinecheck=false
}
\caption{Quantitative results on the Robodrive competition(Track 4).The \textbf{best} scores of each metric are highlighted in \textbf{bold}.}
\label{tab:performance_comparison}
\begin{adjustbox}{width=0.48\textwidth}
\begin{tabular}{@{}lccccccc@{}}
\toprule
Team      & Abs Rel$\downarrow$ & Sq Rel$\downarrow$ & RMSE$\downarrow$  & log RMSE$\downarrow$ & $a1\uparrow$              & $a2\uparrow $            & $a3\uparrow$    \\ \midrule
HIT-AIIA$^1$    & \textbf{0.187}   & \textbf{1.468}  & \textbf{6.236} & \textbf{0.276}    & \textbf{0.734}          & \textbf{0.895}             & \textbf{0.953}            \\
twolones$^2$        & 0.211   & 1.655  & 6.327 & 0.294    & 0.686           & 0.880             & 0.946             \\
CUSTZS$^3$        & 0.264   & 2.320  & 7.961 & 0.363    & 0.578           & 0.816             & 0.913            \\ \bottomrule
\end{tabular}
\end{adjustbox}
\end{table}

\noindent\textbf{Ablation Study}
In Table \ref{tab:ablation_result}, we assess the impact of various repair algorithms, decoders, and backbones. The results demonstrate that all configurations yield improvements in the depth estimation task.

The first line is the official baseline SurroundDepth. We first try DINOv2 as encoder and DPT as decoder. The second line shows that although our DINO-SD do not use any attention mechanism, our DINO-SD performance is better than baseline. In line 3, we try to introduce self attention into DPT and the model performance improved a lot. In lines 4-6, we explore the impact of image denoise and equalization on the results. The results show that both image denoise and equalization help to improve the performance of the model, and the improvement effect is greater when used together. In line 7, we change the self attention into adjacent view cross attention and the results show that the adjacent-view cross attention is better than self attention. The reason adjacent-view cross attention performance better than self attention has explained in S3Depth\cite{S3Depth}, the adjacent views can offer direct environment information compared with non-adjacent views.

\section{Summary}
In this work, we propose the DINO-SD, a novel framework that focuses on improving the robustness of the surround-view depth estimation model. We add attention mechanism into DPT and improved the performance of model. The results show that the proposed method is able to handle various corruptions.

\clearpage
{
\small
\bibliographystyle{ieeenat_fullname}
\bibliography{main}
}

\end{document}